\documentclass[letterpaper, 10 pt, conference]{ieeeconf}  

\IEEEoverridecommandlockouts                              

\usepackage{graphics} 
\usepackage{epsfig} 
\usepackage{times} 
\usepackage{amsmath} 
\usepackage{amssymb}  
\usepackage{cite}
\usepackage{array}
\usepackage{float}
\usepackage{multirow}
\usepackage{booktabs}
\usepackage{subfigure}
\usepackage{algorithm}  
\usepackage{algorithmic}
\title{\LARGE \bf
Coordination and Trajectory Prediction for Vehicle Interactions via Bayesian Generative Modeling
}

\author{Jiachen Li, Hengbo Ma, Wei Zhan and Masayoshi Tomizuka
	\thanks{J. Li, H. Ma, W. Zhan and M. Tomizuka are with the Department of Mechanical Engineering, 
		University of California, Berkeley, CA 94720, USA
		(e-mail: {\tt\small jiachen\_li, hengbo\_ma, wzhan, tomizuka@berkeley.edu})}}

\begin{document}

\maketitle
\thispagestyle{empty}
\pagestyle{empty}

\begin{abstract}
Coordination recognition and subtle pattern prediction of future trajectories play a significant role when modeling interactive behaviors of multiple agents. Due to the essential property of uncertainty in the future evolution, deterministic predictors are not sufficiently safe and robust. 
In order to tackle the task of probabilistic prediction for multiple, interactive entities, we propose a coordination and trajectory prediction system (CTPS), which has a hierarchical structure including a macro-level coordination recognition module and a micro-level subtle pattern prediction module which solves a probabilistic generation task. 
We illustrate two types of representation of the coordination variable: categorized and real-valued, and compare their effects and advantages based on empirical studies. We also bring the ideas of Bayesian deep learning into deep generative models to generate diversified prediction hypotheses. The proposed system is tested on multiple driving datasets in various traffic scenarios, which achieves better performance than baseline approaches in terms of a set of evaluation metrics. The results also show that using categorized coordination can better capture multi-modality and generate more diversified samples than the real-valued coordination, while the latter can generate prediction hypotheses with smaller errors with a sacrifice of sample diversity.
Moreover, employing neural networks with weight uncertainty is able to generate samples with larger variance and diversity. 

\end{abstract}

\section{INTRODUCTION}

There always exists coordination when two or more agents interact with each other. Accurate and timely recognition of coordination outcomes and effective prediction of their future trajectories play a significant role in high-quality and safe decision making and planning. 
In the field of autonomous driving, it is critical for autonomous vehicles to recognize and forecast the behaviors of surrounding vehicles especially when there are potential conflict zones between encountering entities (see Fig. 1). 
In the generalized merging scenarios such as highway ramp merging and roundabout entering scenarios, it is desired to predict the joint distribution of future trajectories of multiple agents when considering interactions.

In this work, a hierarchical system consisting of a macro-level coordination recognition module and a micro-level subtle pattern prediction module is proposed to achieve this goal.
The coordination can be either categories with explicit semantics or real-valued vectors encoding particular traffic situations. If explicit semantics can be clearly defined and labels are available, categorized coordination can be applied. However, if labels are unavailable or too many agents are involved so that it is hard to define categorized coordination explicitly, then only real-valued coordination can be applied.
Subtle pattern denotes the motion patterns which cannot be explicitly categorized via spatial road structure or conflict regions, such as acceleration/deceleration and aggressive/conservative merging.

Deep generative models such as variational auto-encoder (VAE) and generative adversarial network (GAN) have been proved to possess advantages in representation learning and distribution approximation in sample generation tasks. However, there is usually mode collapse issue under the conditional generation settings, especially when they are applied to time-series data. 
In the trajectory prediction tasks, this may lead to very similar prediction hypotheses with low diversity, which may deteriorate the safety and robustness of the prediction system.
To address this problem, we bring in the idea of Bayesian deep learning so that we can incorporate weight uncertainty into the generative neural networks to increase model capacity and sample diversity.

\begin{figure}[!tbp]
	\centering
	\includegraphics[width=\columnwidth,height=7cm]{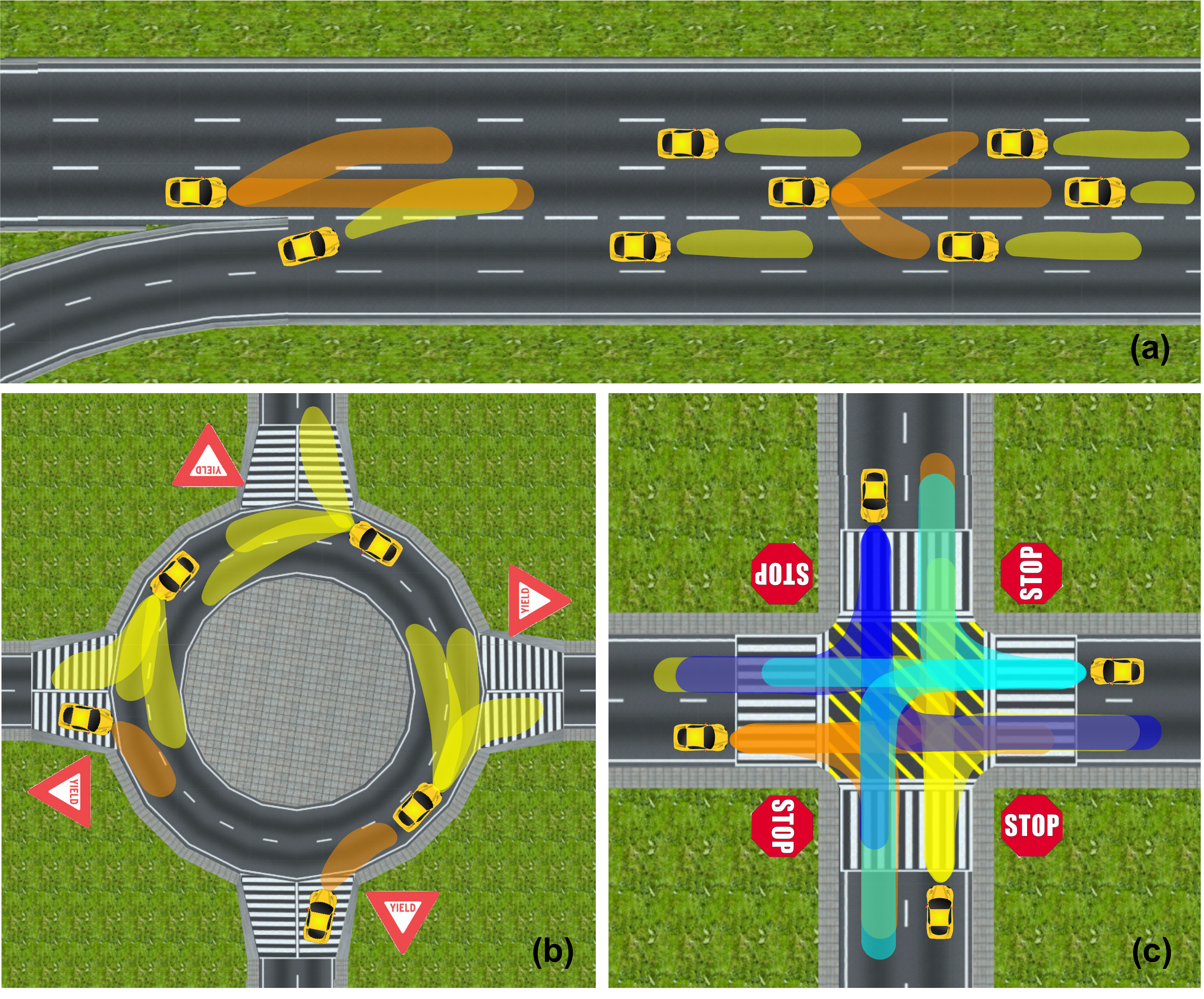}
	\caption{Typical highway and urban driving scenarios where two or more entities coordinate and interact with each other. The shaded areas represent possible future motions which consider multi-modality. (a) Ramp merging and lane change behaviors on highway scenarios; (b) Unsignalized roundabout with yield signs; (c) Unsignalized intersection with stop signs. Although the contexts are different, they can be treated as generalized merging scenarios.}
\end{figure}
The main contributions of this work are summarized as follows:
(a) A Coordination and Trajectory Prediction System (CTPS) is proposed to recognize the coordination outcome of interaction between multiple vehicles and to forecast subtle motion patterns corresponding to a certain coordination.
(b) A latent feature extraction method for time-series data is employed to improve coordination classification accuracy, which is based on a variational recurrent neural network (VRNN).
(c) A Coordination-Bayesian Conditional Generative Adversarial Network (C-BCGAN) is proposed to generate realistic future trajectories given historical information and specific coordination, where weight uncertainties of neural networks are incorporated.

The remainder of the paper is organized as follows. Section II provides a brief overview on related studies. Section III presents a uniform problem formulation for trajectory prediction tasks. Section IV illustrates the proposed methodology. In Section V, the proposed system is applied to several problems of vehicle interaction prediction based on real-world driving datasets. The performance is compared with baseline methods in terms of widely-used evaluation metrics. Finally, Section VI concludes the paper.

\section{Related Work}
In this section, we provide a brief overview on related research and demonstrate the advantages of the proposed prediction system.
\vspace{0.1cm}

\noindent
\textbf{Behavior and Trajectory Prediction}

Many research efforts have been devoted to forecasting behaviors and trajectories of on-road vehicles and pedestrians. 
Many rule-based and model-based approaches were proposed to estimate future states on time-series data, such as variants of Kalman filter (KF) applied to system dynamic models \cite{KF}, auto-regressive models \cite{auto-regressive} and time-series analysis \cite{time-series}. However, these approaches are not able to incorporate the coordination among highly interactive agents due to the limited model flexibility.
Therefore, more learning-based approaches were put forward to tackle complicated scenarios. 
In \cite{traj_pred1}, a variational neural network was employed to predict future trajectories at an urban intersection.
In \cite{traj_pred2}, a dynamic Bayesian network was designed to predict driver maneuvers in highway scenarios.
Other methodologies in the existing literature include hidden Markov models \cite{jiachen_prediction}, Gaussian mixture models \cite{jiachen_tracking}, Gaussian process \cite{traj_pred6} and inverse reinforcement learning \cite{IRL-1,IRL-2}.
In this paper, we propose a hierarchical prediction system based on Bayesian generative modeling and variational inference, which can approximate the data distribution and capture multi-modalities.

\vspace{0.1cm}
\noindent
\textbf{Deep Generative Modeling}

Deep generative models such as variational auto-encoder (VAE) \cite{vae} and generative adversarial networks (GAN) \cite{gan} are widely used in image generation task due to their strengths in approximating distributions of data.
The condition variable can be additionally incorporated to generate samples satisfying particular properties \cite{cgan}.
In this paper, we modify the model framework in \cite{cgan} to generalize the condition into a combination of discrete indicators and real-valued feature vectors.

\vspace{0.1cm}
\noindent
\textbf{Weight Uncertainty in Neural Networks}

The parameters of deep neural networks in most deep learning literature are deterministic, which may not fully represent the model uncertainty. 
In recent years, more attention has been paid to avoid over-fitting or being over-confident on unseen cases by incorporating distributions over network parameters \cite{BGAN,BCGAN}. In \cite{BNN2}, a stochastic dynamical system was modeled by a Bayesian neural network (BNN) and used for model-based reinforcement learning. Moreover, in real-world applications such as autonomous driving, the surrounding environment is highly dynamic and full of uncertainty. \cite{BNN1} suggested employing Bayesian deep learning to address safety issues.
In this work, we treat the parameters in the proposed system as sampled instances from the learned distributions, which improves uncertainty modeling and increases the diversity of prediction hypotheses.

\section{Problem Formulation}
Two or more vehicles will behave coordinately and interactively when conflict zone exists. 
The objective of this paper is to develop a prediction system that can accurately forecast the coordination of 
multiple interactive agents and multi-modal distributions of their future trajectories. 
The system should take into account historical states, context information as well as interactions among dynamic entities.

Assume there are totally $N$ entities in an interactive situation where $N$ may vary in different traffic scenarios. 
We denote a set of possible coordination as $\mathbf{C}$ and two sets of trajectories covering history and prediction horizons as $\mathbf{T}_{k-T_h+1:k}=\{\mathbf{t}^j_{k-T_h+1:k}|\mathbf{t}^j_k=(x^j_k,y^j_k),j=1,...,N\}$ and $\mathbf{T}_{k+1:k+T_f}=\{\mathbf{t}^j_{k+1:k+T_f}|\mathbf{t}^j_k=(x^j_k,y^j_k),j=1,...,N\}$, where $T_h$ and $T_f$ are horizon lengths. 
The $(x,y)$ is the 2D coordinate in the Euclidean world space or image pixel space and $k$ represents the current time step. In this paper, we only focus on the prediction in the Euclidean space.

Mathematically, our goal is to obtain the multi-modal conditional distribution of future trajectories given the historical information $p(\mathbf{T}_{k+1:k+T_f}|\mathbf{T}_{k-T_h+1:k})$. Some necessary features derived from the trajectory information can also be incorporated as condition variables. The long-term prediction can be obtained by propagating the generative system multiple times to the future.

\begin{figure*}[!tbp]
	\centering
	\includegraphics[width=\textwidth,height=8.8cm]{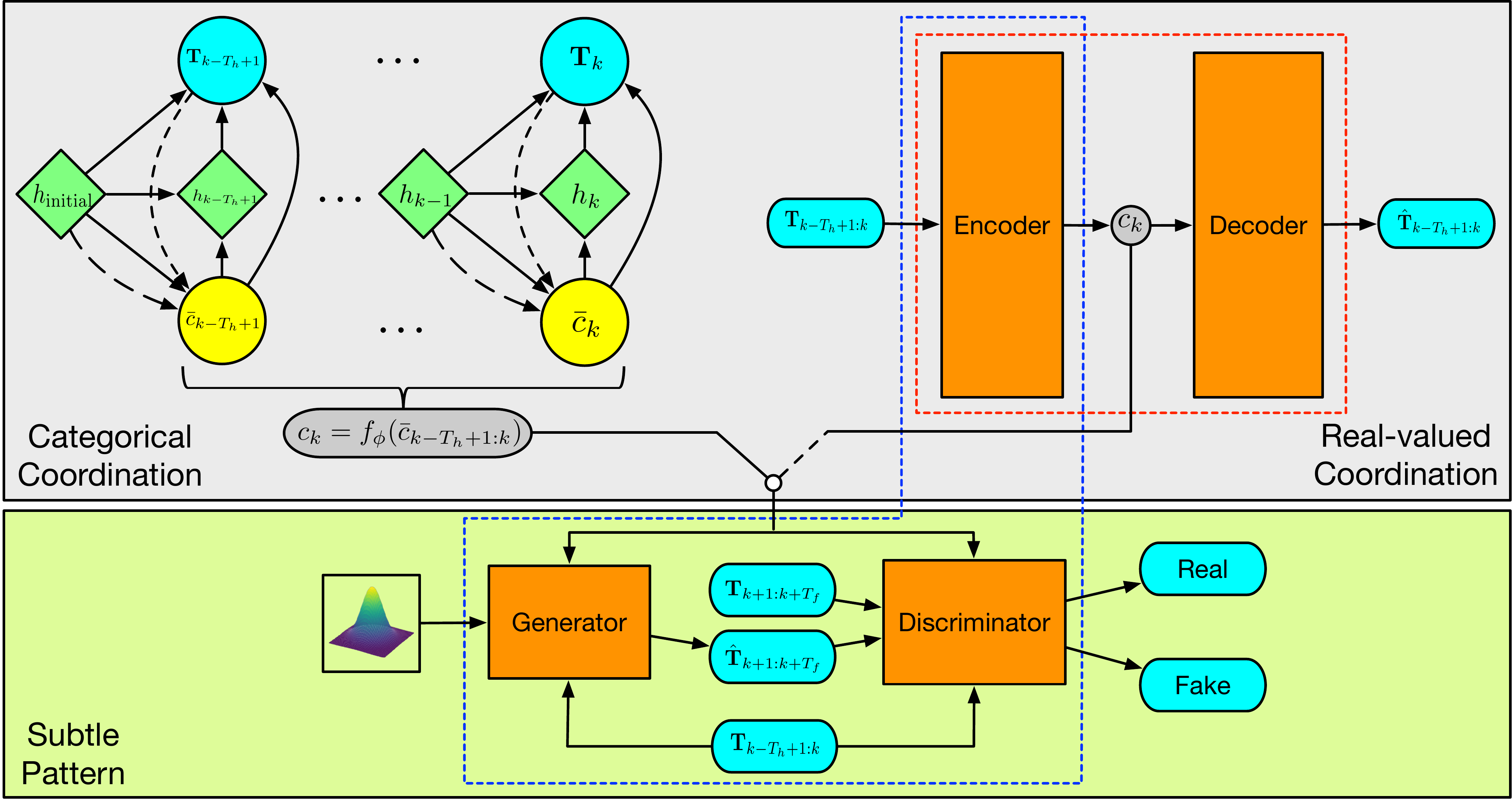}
	\caption{The overview of proposed coordination and trajectory prediction system (CTPS), which consists of two key components: (a) Coordination recognition module: The coordination variable can be discrete categories or continuous real-valued vectors. The discrete distribution of categorized coordination is obtained by a probabilistic classifier based on latent features extracted by VRNN. The continuous distribution of real-valued coordination is obtained by maximizing mutual information based on a VAE-style model. We can choose either formulation according to the objective and emphasis in particular tasks; (b) Subtle pattern prediction module: The model is based on the proposed C-BCGAN in which the generator takes as input the historical information, coordinator indicator as well as a noise from the normal distribution. Weight uncertainties are incorporated in both generator and discriminator network.}
\end{figure*}
\section{Methodology}
In this section, we first provide an overview of the proposed prediction system. Then the details of model components, architecture and algorithms are illustrated.

\subsection{System Overview}
The multi-modal conditional distribution of future trajectories $p(\mathbf{T}_{k+1:k+T_f}|\mathbf{T}_{k-T_h+1:k})$ for interactive agents can be factorized into
\begin{equation*}
\small
\begin{aligned}
&\quad p(\mathbf{T}_{k+1:k+T_f}|\mathbf{T}_{k-T_h+1:k}) = \\ &\sum_{i=1}^{m} p(\mathbf{T}_{k+1:k+T_f}|c_i, \mathbf{T}_{k-T_h+1:k})  p(c_i|\mathbf{T}_{k-T_h+1:k}), \text{if}  \ \text{categorized}, \\
&\int_{c} p(\mathbf{T}_{k+1:k+T_f}|c, \mathbf{T}_{k-T_h+1:k})  p(c|\mathbf{T}_{k-T_h+1:k}), \text{if}  \ \text{real-valued},
\end{aligned}
\end{equation*}
where $m$ is number of possible categorized coordination.
This factorization naturally divides the system into a coordination recognition module (macro-level) and a subtle pattern prediction module (micro-level). The coordination $c$ can not only be categorized to represent meaningful semantics, but also be real-value vectors to encode the underlying representations. If $c$ is categorized, the micro-level module takes $c$ in as an indicator through one-hot encoding; if $c$ is a real-valued variable, the micro-level module takes $c$ in as an additional input feature. The macro-level module is based on a variational recurrent neural network (VRNN) followed by a probabilistic classifier. And the micro-level module is based on a Coordination-Bayesian Conditional Generative Adversarial Network (C-BCGAN). The architecture of the proposed system is demonstrated in Fig. 2. The details of each module are illustrated in the following sections.

\subsection{Coordination Recognition}
The coordination can be either learned with explicit semantics or real-valued vectors encoding the underlying representation of traffic situations. 
The learned coordination can be obtained by solving a probabilistic classification task, and the real-valued coordination can be obtained by fully unsupervised learning.
In this paper, we employ both formulations in the experiments and compare their performance and effects on the subtle pattern prediction.
\vspace{0.1cm}

\noindent
\textbf{Categorized coordination}

The learned coordination model is composed of a VRNN based latent feature extractor and a probabilistic classifier.
VRNN is essentially a recurrent version of variational auto-encoder (VAE) to model time-series data, where the prior distribution of latent variable can evolve along time instead of constant normal distribution. 
We denote $\bar{c}$ as a latent random variable which encodes the input features extracted from historical information. Instead of choosing the normal distribution as the prior of $\bar{c}$ in standard VAE, we assume the following distribution
\begin{equation}
\small
\bar{c}_k \sim \mathcal{N}(\mu_{0,k}, \text{diag}(\sigma^2_{0,k})), \  [\mu_{0,k}, \sigma_{0,k}] = \varphi^{prior}_{\tau}(h_{k-1}),
\end{equation}
where $\mu_{0,k}, \sigma_{0,k}$ denote the mean and variance of the conditional prior distribution. The generation distribution is given as
\begin{equation}
\small
\mathbf{T}_k | \bar{c}_k \sim \mathcal{N}(\mu_{\mathbf{T},k}, \text{diag}(\sigma^2_{\mathbf{T},k})), \ [\mu_{\mathbf{T},k}, \sigma_{\mathbf{T},k}] = \varphi^{dec}_{\tau}(\bar{c}_k,h_{k-1}),
\end{equation}
where $\mu_{\mathbf{T},k}, \sigma_{\mathbf{T},k}$ denote the mean and variance of the generation distribution. 
The hidden state of the RNN is updated by
\begin{equation}
\small
h_k = \varphi^{rnn}_{\tau}(\mathbf{T}_k, \bar{c}_k, h_{k-1}),
\end{equation}
where $\varphi^{rnn}_{\tau}$ is the transition function of recurrent dynamics.
According to the above equations and distributions, we can obtain the joint distribution of historical information $\mathbf{T}_{k-T_h+1:k}$ and latent random variable $\bar{c}_{k-T_h+1:k}$, which is computed by
\begin{equation}
\small
\begin{aligned}
&p(\mathbf{T}_{k-T_h+1:k}, \bar{c}_{k-T_h+1:k}) \\ =& \Pi^{k}_{t=k-T_h+1} p(\mathbf{T}_t|\bar{c}_{\leq t},\mathbf{T}_{<t})p(\bar{c}_t|\mathbf{T}_{<t}, \bar{c}_{<t}).
\end{aligned}
\end{equation}
When conducting inference, the approximated posterior distribution of latent variable at each time step can be obtained by
\begin{equation}
\small
\bar{c}_k | \mathbf{T}_k \sim \mathcal{N}(\mu_{\bar{c},k}, \text{diag}(\sigma^2_{\bar{c},k})), \ [\mu_{\bar{c},k},\sigma_{\bar{c},k}] = \varphi^{enc}_{\tau}(\mathbf{T}_k, h_{k-1}),
\end{equation}
which leads to a factorization as 
\begin{equation}
\small
p(\bar{c}_{k-T_h+1:k}|\mathbf{T}_{k-T_h+1:k}) = \Pi^{k}_{t=k-T_h+1} q(\bar{c}|\mathbf{T}_{\leq t}, \bar{c}_{<t}).
\end{equation}
The objective function of VRNN is formulated as a time-step-wise evidence lower bound (ELBO), which is given by
\begin{equation}
\small
\begin{aligned}
\mathcal{L}_\text{VRNN}=  \mathbb{E}&_{q(\bar{c}_{k-T_h+1:k}|\mathbf{T}_{k-T_h+1:k})} [\sum_{t=k-T_h+1}^{k} \log p(\mathbf{T}_t|\bar{c}_{\leq t},\mathbf{T}_{<t})  \\ - &D_\text{KL}(q(\bar{c}_t|\mathbf{T}_{<t}, \bar{c}_{<t})||p(\bar{c}_t|\mathbf{T}_{<t}, \bar{c}_{<t}))],
\end{aligned}
\end{equation}
where $D_\text{KL}$ is the Kullback-Leibler divergence.
The aforementioned functions $\varphi^{prior}_{\tau},\varphi^{dec}_{\tau},\varphi^{enc}_{\tau},\varphi^{rnn}_{\tau}$ can be approximated by neural networks.

Finally, we apply a classifier to the latent variable sequence to obtain the final coordination distribution,
\begin{equation}
\small
p(c_k|\mathbf{T}_{k-T_h+1:k}) = f_{\phi}(\bar{c}_{k-T_h+1:k}),
\end{equation}
where $f_{\phi}$ can be any type of probabilistic classifiers. In this work, we employ a fully connected neural network followed by a softmax layer.
\vspace{0.1cm}

\noindent
\textbf{Real-valued coordination}

In order to obtain real-valued coordination by fully unsupervised learning, we employ a model similar to standard VAE to encode the entire historical information sequence into a single latent coordination variable $c_k$.
To enhance the mutual dependence, we also maximize the mutual information between $c_k$ and historical information $\mathbf{T}_{k-T_h+1:k}$.
The ELBO in this case is formulated as
\begin{equation}
\small
\begin{aligned}
\mathcal{L}_1 = \mathbb{E}&_{q_\phi(c_k|\mathbf{T}_{k-T_h+1:k})} [ \log p_{\theta}(\mathbf{T}_{k-T_h+1:k}|c_k) \\
&- D_{\text{KL}}(q_\phi(c_k|\mathbf{T}_{k-T_h+1:k})||p_\theta(c_k))],
\end{aligned}
\end{equation}
where $q_\phi(c_k|\mathbf{T}_{k-T_h+1:k})$, $p_{\theta}(\mathbf{T}_{k-T_h+1:k}|c_k)$ and $p_\theta(c_k)$ represent the encoder network, decoder network and prior distribution (normal distribution), respectively.
Since it is intractable to obtain the exact mutual information between $\mathbf{T}_{k-T_h+1:k}$ and $c_k$, we need a variational distribution $q_\psi(c_k|\mathbf{T}_{k-T_h+1:k})$ to derive a variational lower-bound of mutual information (MI)
\begin{equation}
\small
\begin{aligned}
\mathcal{L}_2 &= H(c_k) + \mathop{\mathbb{E}}\limits_{p_\theta(\mathbf{T}_{k-T_h+1:k}|c_k)} \left[ \mathbb{E}_{q_\phi}[\log q_\psi(c_k|\mathbf{T}_{k-T_h+1:k})]\right] \\ &\leq MI(\mathbf{T}_{k-T_h+1:k},c_k),
\end{aligned}
\end{equation}
where $H(c_k)=-\int_{c_k} p(c_k)\log p(c_k) = -\mathbb{E}_{p(c_k)}\log p(c_k)$. The model parameters are updated by maximizing $\mathcal{L}_1 + \lambda \mathcal{L}_2$, where $\lambda$ is chosen to be unit in this work.

\subsection{Subtle Pattern Prediction}
In this module, we propose an Coordination-Bayesian Conditional GAN (C-BCGAN) as the basis of subtle pattern prediction, which transforms a random noise $z$ into the future trajectory distribution conditioned on historical information $\mathbf{T}_{k-T_h+1:k}$ and coordination $c_k$.
The goal is to estimate $p(\mathbf{T}_{k+1:k+T_f}|c^i_k, \mathbf{T}_{k-T_h+1:k})$ for each learned coordination and $p(\mathbf{T}_{k+1:k+T_f}|c_k, \mathbf{T}_{k-T_h+1:k})$ for real-valued coordination.

The proposed C-BCGAN consists of a generator network $G(z|c_k, \mathbf{T}_{k-T_h+1:k}; \theta_g)$ to produce prediction hypotheses from the learned data distribution and a discriminator network $D(\mathbf{T}_{k+1:k+T_f}|c_k, \mathbf{T}_{k-T_h+1:k}; \theta_d)$ to distinguish whether the samples come from the real data distribution or the learned generation distribution.

Instead of making point mass estimation on the weight of neural networks in standard GAN methods, we introduce weight uncertainty by placing distributions over $\theta_g$ and $\theta_d$ to increase the diversity of generated samples as well as to alleviate the overfitting and mode collapse problems in the training process.
In order to obtain the posterior distribution over network weights, samples are drawn iteratively according to the following conditional distributions:
\begin{equation}
\small
\begin{aligned}
&p(\theta_g|z,c_k, \mathbf{T}_{k-T_h+1:k}, \theta_d) \\ \propto& \left(\mathop{\Pi}_{i=1}^{n_g} D(G(z^{(i)}|c_k, \mathbf{T}_{k-T_h+1:k}))\right)p(\theta_g),
\end{aligned}
\end{equation}
\begin{equation}
\small
\begin{aligned}
p(\theta_d|z,&X,c_k, \mathbf{T}_{k-T_h+1:k},\theta_g) \propto \mathop{\Pi}_{i=1}^{n_d} D(x^{(i)}|c_k, \mathbf{T}_{k-T_h+1:k}) \\
&\times \mathop{\Pi}_{i=1}^{n_g} (1-D(G(z^{(i)}|c_k, \mathbf{T}_{k-T_h+1:k}))) p(\theta_d),
\end{aligned}
\end{equation}
where $p(\theta_g)$, $p(\theta_d)$ are the prior distributions, and $n_g$, $n_d$ are the batch sizes of generator and discriminator, respectively. $X$ represents a batch of $n_d$ samples drawn from the real data distribution.

The marginalized posterior distribution can be calculated by
\begin{equation}
\small
\begin{aligned}
&p(\theta_g|c_k,\mathbf{T}_{k-T_h+1:k}, \theta_d) = \int p(\theta_g,z|c_k,\mathbf{T}_{k-T_h+1:k},\theta_d)\text{d}z \\
=& \int p(\theta_g|z,c_k,\mathbf{T}_{k-T_h+1:k},\theta_d)p(z|c_k,\mathbf{T}_{k-T_h+1:k},\theta_d)\text{d}z \\
\approx& \frac{1}{J_g} \sum_{j=1}^{J_g} p(\theta_g|z^{(j)},c_k,\mathbf{T}_{k-T_h+1:k},\theta_d),
\end{aligned}
\end{equation}
\begin{equation}
\small
\begin{aligned}
&p(\theta_d|c_k,\mathbf{T}_{k-T_h+1:k}, X, \theta_g) = \int p(\theta_d,z|c_k,\mathbf{T}_{k-T_h+1:k},X,\theta_g)\text{d}z \\
=& \int p(\theta_d|z,c_k,\mathbf{T}_{k-T_h+1:k},X,\theta_g)p(z|c_k,\mathbf{T}_{k-T_h+1:k},X,\theta_d)\text{d}z \\
\approx& \frac{1}{J_d} \sum_{j=1}^{J_d} p(\theta_d|z^{(j)},c_k,\mathbf{T}_{k-T_h+1:k},X,\theta_g),
\end{aligned}
\end{equation}
from which the network weights can be sampled.
To sample from the posterior distribution, we employ the Stochastic Gradient Hamiltonian Monte Carlo (SGHMC) method \cite{SGHMC} which is illustrated in Algorithm 1.
In order to speed up convergence to reasonable prediction hypotheses in the early stage, we add $L2$ norm of the difference between prediction hypotheses and groundtruth as well as differentiable barrier functions into the loss function, which can incorporate soft constraints on the prediction output \cite{CPN}.
\begin{algorithm*}[!tbp]
	\caption{The SGHMC algorithm for sampling weights of generator and discriminator from posterior distributions, which is modified from \cite{BGAN}. $\alpha$ is the friction term and $\eta$ is the learning rate. $J_g$ and $J_d$ MC samples are drawn for the generator and discriminator and $M$ SGHMC samples for each MC samples. The posterior distributions are represented by sample sets $\{\theta_g^{j,m}\}^{J_g,M}_{j=1,m=1}$ and $\{\theta_d^{j,m}\}^{J_d,M}_{j=1,m=1}$ at the last iteration.}
	\begin{algorithmic}[1]
		\FOR {$J_g$}
		\STATE Sample $J_g$ noise values $\{z^{(1)},...,z^{(J_g)}\}$ from prior $p(z)$. Each $z^{(i)}$ has $n_g$ samples.\\
		\STATE Update sample set representing $p(\theta_g|c,\mathbf{T}_{k-T_h:k}, \theta_d)$ by SGHMC updates for $M$ iterations:\\
		\STATE \begin{equation}
		\theta_g^{j,m}\gets \theta_g^{j,m}+v; v\gets (1-\alpha)v+\eta \frac{\partial \log \left(\sum_{i}\sum_{k}p(\theta_g|z^{(j)},c,\mathbf{T}_{k-T_h:k},\theta_d)\right)}{\partial \theta_g}+n; n \sim \mathcal{N}(0, 2\alpha\eta I);
		\end{equation}
		\STATE Append $\theta_g^{j,m}$ to the sample set.
		\ENDFOR
		\FOR {$J_d$}
		\STATE Sample $J_d$ noise values $\{z^{(1)},...,z^{(J_d)}\}$ from prior $p(z)$;
		\STATE Sample $n_d$ data samples $x$;
		\STATE Update sample set representing $p(\theta_d|c,\mathbf{T}_{k-T_h:k}, X, \theta_g)$ by SGHMC updates for $M$ iterations:\\
		\STATE  \begin{equation}
		\theta_d^{j,m}\gets \theta_d^{j,m}+v; v\gets (1-\alpha)v+\eta \frac{\partial \log \left(\sum_{i}\sum_{k}p(\theta_d|z^{(j)},c,\mathbf{T}_{k-T_h:k},X,\theta_g)\right)}{\partial \theta_d}+n; n \sim \mathcal{N}(0, 2\alpha\eta I);
		\end{equation}
		\STATE Append $\theta_d^{j,m}$ to the sample set.
		\ENDFOR
	\end{algorithmic}
\end{algorithm*}
\subsection{Coordination and Trajectory Prediction System (CTPS)}
We combine the coordination recognition module and subtle pattern prediction module hierarchically to obtain the coordination and trajectory prediction system.
The generator takes in coordination, latent noise and historical information and outputs prediction hypotheses. The discriminator takes in coordination, historical information, groundtruth and prediction hypotheses and outputs the probability that the samples come from the real data distribution.
\vspace{0.1cm}

\noindent
\textbf{Categorized Coordination}

In the training phase, we first train the VRNN with unsupervised learning to extract latent features, which are employed as training features for probabilistic classifiers.
The coordination is one-hot-encoded as an input feature of generator and discriminator.
Then the subtle pattern prediction module is trained by adversarial learning.
In the testing phase, we first obtain the coordination probability. Then we sample a set of future trajectories corresponding to each coordination. The number of samples of each class is proportional to the class probability.

\vspace{0.1cm}

\noindent
\textbf{Real-valued Coordination}

In the training phase, we first train the coordination model by unsupervised learning to obtain an optimal encoder under the condition of mutual information maximization according to the procedures in section IV-B.
Then, by connecting the trained encoder to the C-BCGAN we jointly train the encoder, generator and discriminator to reach a Nash equilibrium. These two procedures are iterated until all the parameters converge to a local optimum. 
In the testing phase, the coordination variable is sampled by the encoder and passed to C-BCGAN to generate trajectory hypotheses.

\section{Experiments and Discussion}
In this section, we validate the proposed approach on two dataset of naturalistic driving, which covers vehicle interactions in ramp merging, standard highway and roundabout scenarios. The model performance is compared with several state-of-the-art baselines in terms of a set of evaluation metrics.

\subsection{Datasets and Scenarios}
In order to demonstrate the effectiveness and generality of the proposed approach, we evaluate it on multiple naturalistic driving datasets covering various typical scenarios.

(a) \textit{NGSIM}: The NGSIM US-101 highway dataset \cite{NGSIM} was used to extract training and testing data. For the ramp merging scenario, we focus on the coordination and interaction between two vehicles, the one on the ramp and the one on the main lane who shares a conflict zone. The learned coordination contains \textit{pass} and \textit{yield} (merging vehicle).
For the lane change and lane keeping behaviors, we focus on the coordination of six vehicles where the one at the center may change lanes but the others are assumed to keep their lanes, which is similar to the formulation in \cite{jiachen_prediction_gan}. The learned coordination contains \textit{lane change left}, \textit{lane keeping} and \textit{lane change right}. 

(b) \textit{Roundabout}: The raw dataset was collected by our drone equipped with high-resolution cameras. The vehicles were detected by deep neural networks and the trajectories were extracted by filtering the centers of bounding boxes. The learned coordination contains \textit{pass} and \textit{yield} (entering vehicle).

\subsection{Baseline Methods}
We compare the performance of our proposed approach with the following baseline methods on the aforementioned datasets. The coordination recognition and subtle pattern prediction are evaluated separately.

(a) \textit{Hidden Markov Model (HMM)}: The HMM is a widely used probabilistic classifier, which can be employed to obtain the probability of each learned coordination.

(b) \textit{Gaussian Naive Bayes (GNB)}: NB is a typical probabilistic
classifier which employs Bayes’ theorem and assumes that features are mutually independent. In this work, we assume that the feature likelihood to be Gaussian distribution, which establishes a GNB.

(c) \textit{Probabilistic GRU (P-GRU)}: A vanilla recurrent neural network with GRU cells. A noise term sampled from the normal distribution is appended in the input features to incorporate uncertainty.

(d) \textit{Conditional Variational Auto-encoder (CVAE)}: The model is modified from \cite{CVAE} which is generalized to time-series generation. The encoder and decoder both take in the historical information in the training phase. In the test phase, we only use the decoder to generate prediction hypotheses.

(e) \textit{Conditional Generative Adversarial Network (CGAN)}: The model is modified from \cite{cgan} which is also generalized to time-series generation. The training and testing processes are similar to the original method.

\subsection{Evaluation Metrics}
The proposed coordination recognition and subtle pattern prediction methods are evaluated from the following aspects, respectively. 
We denote the batch size as $B$ and the number of generated samples as $N_s$.

(a) \textit{Precision and Recall}: These metrics are employed to evaluate the performance of learned coordination recognition in this paper, which are calculated by
\begin{equation}
\small
\begin{aligned}
precision &= \frac{TP}{TP+FP}, \ recall = \frac{TP}{TP+FN}, \\
&F1 = \frac{2*precision* recall}{precision+recall},
\end{aligned}
\end{equation}
where $TP,FP,FN$ represent true positives, false positives and false negatives, respectively. More details can be found in \cite{precision_recall}.

(b) \textit{Diversity:} The step-wise Euclidean distance based on the mean square error (MSE) between each pair of the generated trajectories is calculated to measure the degree of sample diversity, which is calculated as
\begin{equation}
\small
Diversity  = \sqrt{\frac{1}{B(N_s-1)} \sum_{b=1}^{B}\sum_{i=1}^{N_s} \sum_{j\neq i} ||t^{b,i}_{1:T_f} - t^{b,j}_{1:T_f}||^2}
\end{equation}

(c) \textit{Minimum Distance} ($Dist_{\min}$): We evaluate the quality of generations by measuring the minimum Euclidean distance of the closest sample among all the prediction hypotheses to the ground truth. This metric stresses the highest prediction capability in terms of accuracy.
\begin{equation}
\small
Dist_{\min} = \sqrt{\frac{1}{B} \sum_{b=1}^{B}\min_i ||t^{b,i}_{1:T_f}-t^{b,gt}_{1:T_f}||^2}
\end{equation}

(d) \textit{Average Distance} ($Dist_{\text{avg}}$): We also evaluate the quality of generations by measuring the average Euclidean distance of all the samples to the ground truth. This metric stresses the average prediction capability.
\begin{equation}
\small
Dist_{\text{avg}} = \sqrt{\frac{1}{BN_s} \sum_{b=1}^{B}||t^{b,i}_{1:T_f}-t^{b,gt}_{1:T_f}||^2}
\end{equation}

(e) \textit{Final Distance} ($Dist_{\text{final}}$): The final output of a reasonable and interpretable approach should be bounded in a reasonable space around the ground truth. Therefore, we compare the distance at the final time step, which is given by
\begin{equation}
Dist_{\text{final}} = \sqrt{\frac{1}{BN_s} \sum_{b=1}^{B}||t^{b,i}_{T_f}-t^{b,gt}_{T_f}||^2}
\end{equation}

\subsection{Implementation Details}
For the ramp merging and highway scenarios, we predict the future trajectories of 4s given the historical information of 1s. 
For the roundabout scenario, we only forecast future trajectories of 2s since it is long enough to tell the coordination outcomes. 
The same model architecture is applied to all the scenarios. 
Specifically, the $\varphi^{enc}_{\tau}$, $\varphi^{dec}_{\tau}$ and $\varphi^{prior}_{\tau}$ of VRNN are all three-layer MLP and 
the $\varphi^{rnn}_{\tau}$ consists of GRU cells with 64 hidden units. 
The generator and discriminator both consist of GRU cells with 128 hidden units. 
The encoder and decoder both are three-layer MLP with 128 hidden units.
We set the dimension of real-valued coordination vector as three.
The hyper-parameters in Algorithm 1 are as follows: $\alpha=0.9, \eta=0.001, J_g=J_d=10 $ and $M=10$. 

\subsection{Quantitative Analysis}

\begin{table*}[!tbp]
	\caption{Precision / Recall / F1 scores for Categorized Coordination Recognition}
	\label{tab:feature}
	\begin{center}
		\begin{tabular}{m{1.9cm}<{\centering}| m{1.6cm}<{\centering}| m{1.6cm}<{\centering} | m{1.6cm}<{\centering} | m{1.6cm}<{\centering} | m{1.6cm}<{\centering} | m{1.6cm}<{\centering}}
			\toprule
			\midrule
			&  HMM & GNB & MLP & VRNN+HMM & VRNN+GNB & VRNN+MLP \\ 
			\midrule
			Ramp Merging  & 0.66/0.78/0.72 & 0.72/0.74/0.73 & 0.86/0.90/0.88 & 0.88/0.92/0.90   &   0.78/0.84/0.81   &   \textbf{0.91}/\textbf{0.93}/\textbf{0.92}    \\
			Highway       & 0.73/0.79/0.76 & 0.72/0.80/0.76 & 0.85/0.90/0.87 & 0.76/0.86/0.81   &   0.77/0.83/0.80   &   \textbf{0.87}/\textbf{0.91}/\textbf{0.89}    \\
			Roundabout    & 0.86/0.88/0.87 & 0.85/0.86/0.85 & 0.89/0.86/0.87 & \textbf{0.92}/0.87/0.89   &   0.90/0.88/0.89   &   0.91/\textbf{0.90}/\textbf{0.90}    \\		
			\bottomrule
		\end{tabular}
	\end{center}
\end{table*}

\begin{table*}[!tbp]
	\caption{Evaluation of Subtle Pattern Prediction (Vehicle Position)}
	\label{tab:feature}
	\begin{center}
		\begin{tabular}{m{2cm}<{\centering}| m{2cm}<{\centering}| m{2cm}<{\centering}| m{2cm}<{\centering}| m{2cm}<{\centering}| m{2cm}<{\centering}| m{2cm}<{\centering} m{2cm}<{\centering}}
			\toprule
			\midrule
			Scenarios & Evaluation Metrics (m) & CTPS (Categorized) & CTPS (Real-valued) &P-GRU & CVAE & CGAN  \\ 
			\midrule
			\multirow{4}*{\shortstack[lb]{Ramp Merging}}  
			&Diversity & \textbf{0.189} & 0.111 & 0.135 & 0.098 & 0.073\\
			&$Dist_{\min}$ & 0.103 & \textbf{0.087} & 0.268 & 0.134 & 0.156\\
			&$Dist_{\text{avg}}$ & 0.242 & \textbf{0.186} & 0.276 & 0.261 & 0.293\\
			&$Dist_{\text{final}}$ & 0.367 & \textbf{0.232} & 0.313 & 0.362 & 0.351\\
			\midrule
			\multirow{4}*{\shortstack[lb]{Highway}}  
			&Diversity & \textbf{4.512} & 3.112 & 2.192 & 2.576 & 1.684\\
			&$Dist_{\min}$ & \textbf{0.576} & 0.710 & 0.648 & 0.646 & 0.588\\
			&$Dist_{\text{avg}}$ & 1.724 & \textbf{1.042} & 1.466 & 1.136 & 1.226\\
			&$Dist_{\text{final}}$ & 3.266 & \textbf{1.854} & 2.964 & 1.784 & 2.524\\
			\midrule
			\multirow{4}*{\shortstack[lb]{Roundabout}}  
			&Diversity & \textbf{0.213} & 0.188 & 0.126 & 0.167 & 0.112\\
			&$Dist_{\min}$ & 0.048 & 0.043 & 0.061 & \textbf{0.041} & 0.075\\
			&$Dist_{\text{avg}}$ & 0.151 & \textbf{0.112} & 0.186 & 0.143 & 0.137\\
			&$Dist_{\text{final}}$ & 0.207 & 0.196 & 0.231 & 0.191 & \textbf{0.182}\\
			\bottomrule
		\end{tabular}
	\end{center}
\end{table*}
\begin{figure*}[!tbp]
	\centering
	\includegraphics[width=\textwidth,height=5.4cm]{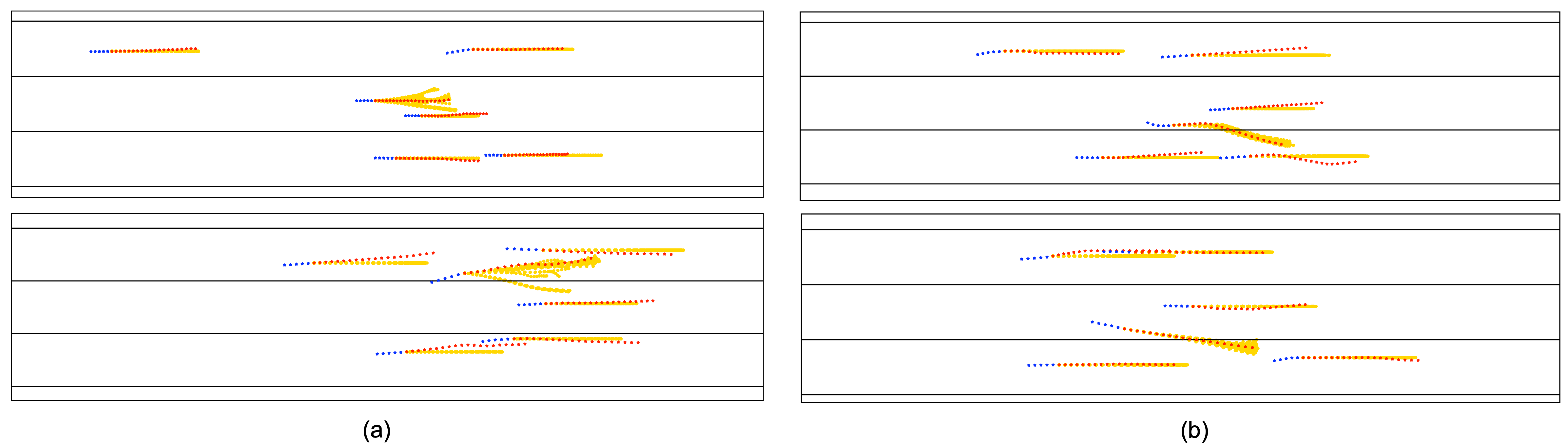}
	\caption{The visualization of prediction results in the highway scenario. (a) Generation with learned coordination; (b) Generation with real-valued coordination. Note that we only predict the longitudinal motions for surrounding vehicles but both longitudinal and lateral motions for the center vehicle. That is the reason why the predicted trajectories of surrounding vehicles do not have lateral deviation.}
\end{figure*}
\begin{figure*}[!tbp]
	\centering
	\includegraphics[width=0.7\textwidth]{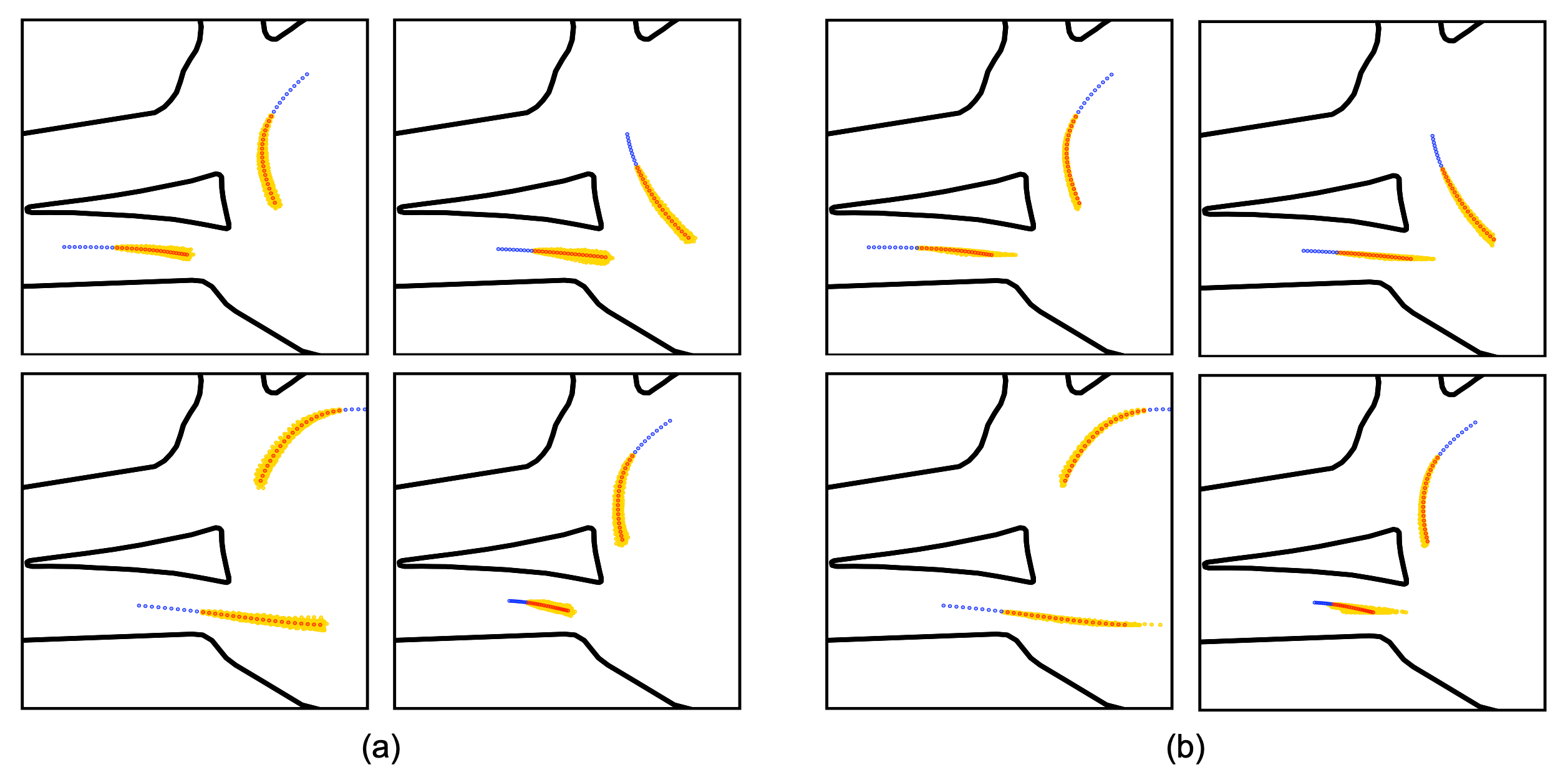}
	\caption{The visualization of prediction results in the roundabout scenario. (a) Generation by C-BCGAN; (b) Generation by C-CGAN (without weight uncertainty). The test cases in (a) and (b) are identical to illustrate the effect of incorporating weight uncertainty.}
\end{figure*}
Since the learned coordination recognition module essentially solves a classification task, its performance in terms of precision, recall and $F1$ score are shown in Table I. The bold numbers represent the best performance among the compared methods.
In order to illustrate the effect of latent feature extraction, we applied different classifiers to both the original features and the latent features extracted by VRNN and compared the classification accuracy and robustness.
It can be seen that with the same type of classifier, the classification results based on latent features achieves higher scores, which implies that the encoded latent feature sequences have a more distinguishable pattern for different coordination categories than the original historical trajectories.
Moreover, MLP can achieve the best performance compared with traditional statistical classifiers because of its higher model capacity.

We also made a comparison on the aforementioned evaluation metrics for subtle pattern prediction module, which is shown in Table II.  
It is shown that the proposed CTPS with learned coordination can generate samples with the largest diversity regardless of datasets and scenarios. The reason is that the learned coordination imposes a multi-modal nature on the generation process which the other methods can not necessarily capture.
Moreover, incorporating uncertainty of generator parameters leads to larger variance of generated samples.
In terms of average prediction accuracy, the CTPS with real-valued coordination achieves the best performance, which implies a tradeoff between diversity (multi-modality) and accuracy.

\subsection{Qualitative Analysis}

We visualize the prediction results of typical cases in highway and roundabout scenarios to illustrate the effects of coordination types and weight uncertainty of networks. 
In Fig. 3(a), we can see a clear multi-modality by using learned coordination, which implies that we can indeed manipulate the generated samples by changing the one-hot-encoded indicator. 
In Fig. 3(b), the real-valued coordination is able to capture the true outcome but the generated trajectories by sampling from the latent space are very similar, where multi-modality can hardly be captured. A practical reason is the sparsity of training data points in a high dimensional space.
Moreover, in Fig. 4 we illustrate the effects of weight uncertainty on the prediction results under the same historical information.
We can see an obvious larger variance in the prediction samples generated by C-BCGAN than those generated by C-CGAN, which alleviates the mode collapse issue.

\section{CONCLUSIONS}
In this paper, we proposed a coordination and trajectory prediction system (CTPS) for vehicle interactions in generalized merging scenarios, which consists of a coordination recognition module and a subtle pattern prediction module. The coordination distribution can be either a discrete distribution over semantic meanings obtained by a probabilistic classifier based on latent features extracted by VRNN, or a continuous distribution over real-valued latent variables obtained by maximizing mutual information based on a VAE-style model. 
In order to generate diversified trajectories corresponding to different coordination outcomes, we proposed a C-BCGAN to incorporate uncertainty for the weights of neural networks. 
The proposed system has been validated on two driving datasets under different scenarios. The results show that the proposed system achieves better performance than baseline approaches in terms of a set of evaluation metrics.



\section*{ACKNOWLEDGMENT}
The authors thank Liting Sun and Di Wang for their assistance on the data collection and pre-processing of the roundabout dataset.

\bibliographystyle{IEEEtran}
\bibliography{reference}

\begin{thebibliography}{10}
\providecommand{\url}[1]{#1}
\csname url@samestyle\endcsname
\providecommand{\newblock}{\relax}
\providecommand{\bibinfo}[2]{#2}
\providecommand{\BIBentrySTDinterwordspacing}{\spaceskip=0pt\relax}
\providecommand{\BIBentryALTinterwordstretchfactor}{4}
\providecommand{\BIBentryALTinterwordspacing}{\spaceskip=\fontdimen2\font plus
\BIBentryALTinterwordstretchfactor\fontdimen3\font minus
  \fontdimen4\font\relax}
\providecommand{\BIBforeignlanguage}[2]{{%
\expandafter\ifx\csname l@#1\endcsname\relax
\typeout{** WARNING: IEEEtran.bst: No hyphenation pattern has been}%
\typeout{** loaded for the language `#1'. Using the pattern for}%
\typeout{** the default language instead.}%
\else
\language=\csname l@#1\endcsname
\fi
#2}}
\providecommand{\BIBdecl}{\relax}
\BIBdecl

\bibitem{KF}
W.~Liu, H.~He, and F.~Sun, ``Vehicle state estimation based on minimum model
  error criterion combining with extended kalman filter,'' \emph{Journal of the
  Franklin Institute}, vol. 353, no.~4, pp. 834--856, 2016.

\bibitem{auto-regressive}
H.~Akaike, ``Fitting autoregressive models for prediction,'' \emph{Annals of
  the institute of Statistical Mathematics}, vol.~21, no.~1, pp. 243--247,
  1969.

\bibitem{time-series}
H.~Kantz and T.~Schreiber, \emph{Nonlinear time series analysis}.\hskip 1em
  plus 0.5em minus 0.4em\relax Cambridge university press, 2004, vol.~7.

\bibitem{traj_pred1}
X.~Huang, S.~McGill, B.~C. Williams, L.~Fletcher, and G.~Rosman,
  ``Uncertainty-aware driver trajectory prediction at urban intersections,''
  \emph{arXiv preprint arXiv:1901.05105}, 2019.

\bibitem{traj_pred2}
J.~Li, B.~Dai, X.~Li, X.~Xu, and D.~Liu, ``A dynamic bayesian network for
  vehicle maneuver prediction in highway driving scenarios: Framework and
  verification,'' \emph{Electronics}, vol.~8, no.~1, p.~40, 2019.

\bibitem{jiachen_prediction}
J.~Li, H.~Ma, W.~Zhan, and M.~Tomizuka, ``Generic probabilistic interactive
  situation recognition and prediction: From virtual to real,'' in \emph{2018
  IEEE Intelligent Transportation Systems Conference}, 2018.

\bibitem{jiachen_tracking}
J.~Li, W.~Zhan, and M.~Tomizuka, ``Generic vehicle tracking framework capable
  of handling occlusions based on modified mixture particle filter,'' in
  \emph{Proceedings of 2018 IEEE Intelligent Vehicles Symposium (IV)}, 2018,
  pp. 936--942.

\bibitem{traj_pred6}
C.~Laugier, I.~Paromtchik, M.~Perrollaz, Y.~Mao, J.-D. Yoder, C.~Tay,
  K.~Mekhnacha, and A.~N{\`e}gre, ``Probabilistic analysis of dynamic scenes
  and collision risk assessment to improve driving safety,'' \emph{Its
  Journal}, vol.~3, no.~4, pp. 4--19, 2011.

\bibitem{IRL-1}
L.~Sun, W.~Zhan, and M.~Tomizuka, ``Probabilistic prediction of interactive
  driving behavior via hierarchical inverse reinforcement learning,'' in
  \emph{2018 21st International Conference on Intelligent Transportation
  Systems (ITSC)}.\hskip 1em plus 0.5em minus 0.4em\relax IEEE, 2018, pp.
  2111--2117.

\bibitem{IRL-2}
L.~Sun, W.~Zhan, M.~Tomizuka, and A.~D. Dragan, ``Courteous autonomous cars,''
  in \emph{2018 IEEE/RSJ International Conference on Intelligent Robots and
  Systems (IROS)}.\hskip 1em plus 0.5em minus 0.4em\relax IEEE, 2018, pp.
  663--670.

\bibitem{vae}
D.~P. Kingma and M.~Welling, ``Auto-encoding variational bayes,'' \emph{arXiv
  preprint arXiv:1312.6114}, 2013.

\bibitem{gan}
I.~Goodfellow, J.~Pouget-Abadie, M.~Mirza, B.~Xu, D.~Warde-Farley, S.~Ozair,
  A.~Courville, and Y.~Bengio, ``Generative adversarial nets,'' in
  \emph{Advances in neural information processing systems}, 2014, pp.
  2672--2680.

\bibitem{cgan}
M.~Mirza and S.~Osindero, ``Conditional generative adversarial nets,''
  \emph{arXiv preprint arXiv:1411.1784}, 2014.

\bibitem{BGAN}
Y.~Saatci and A.~G. Wilson, ``Bayesian gan,'' in \emph{Advances in neural
  information processing systems}, 2017, pp. 3622--3631.

\bibitem{BCGAN}
M.~E. Abbasnejad, Q.~Shi, I.~Abbasnejad, A.~v.~d. Hengel, and A.~Dick,
  ``Bayesian conditional generative adverserial networks,'' \emph{arXiv
  preprint arXiv:1706.05477}, 2017.

\bibitem{BNN2}
S.~Depeweg, J.~M. Hern{\'a}ndez-Lobato, F.~Doshi-Velez, and S.~Udluft,
  ``Learning and policy search in stochastic dynamical systems with bayesian
  neural networks,'' \emph{arXiv preprint arXiv:1605.07127}, 2016.

\bibitem{BNN1}
R.~McAllister, Y.~Gal, A.~Kendall, M.~Van Der~Wilk, A.~Shah, R.~Cipolla, and
  A.~V. Weller, ``Concrete problems for autonomous vehicle safety: Advantages
  of bayesian deep learning.''\hskip 1em plus 0.5em minus 0.4em\relax
  International Joint Conferences on Artificial Intelligence, Inc., 2017.

\bibitem{SGHMC}
T.~Chen, E.~Fox, and C.~Guestrin, ``Stochastic gradient hamiltonian monte
  carlo,'' in \emph{International Conference on Machine Learning}, 2014, pp.
  1683--1691.

\bibitem{CPN}
W.~Zhan, J.~Li, Y.~Hu, and M.~Tomizuka, ``Safe and feasible motion generation
  for autonomous driving via constrained policy net,'' in \emph{Industrial
  Electronics Society, IECON 2017-43rd Annual Conference of the IEEE}.\hskip
  1em plus 0.5em minus 0.4em\relax IEEE, 2017, pp. 4588--4593.

\bibitem{NGSIM}
J.~Colyar and J.~Halkias, ``{US} highway 101 dataset,'' in \emph{Federal
  Highway Administration (FHWA) Technique Report. FHWA-HRT-07-030}, 2007.

\bibitem{jiachen_prediction_gan}
J.~Li, H.~Ma, and M.~Tomizuka, ``Interaction-aware multi-agent tracking and
  probabilistic behavior prediction via adversarial learning,'' in \emph{2018
  IEEE International Conference on Robotics and Automation (ICRA)}, 2019.

\bibitem{CVAE}
K.~Sohn, H.~Lee, and X.~Yan, ``Learning structured output representation using
  deep conditional generative models,'' in \emph{Advances in Neural Information
  Processing Systems}, 2015, pp. 3483--3491.

\bibitem{precision_recall}
D.~M. Powers, ``Evaluation: from precision, recall and f-measure to roc,
  informedness, markedness and correlation,'' 2011.

\end{thebibliography}

\end{document}